\documentclass{article}

\usepackage[utf8]{inputenc} 
\usepackage[T1]{fontenc}    
\usepackage{hyperref}       
\usepackage{url}            
\usepackage{booktabs}       
\usepackage{amsfonts}       
\usepackage{nicefrac}       
\usepackage{microtype}      
\usepackage{xcolor}         

\usepackage{graphicx}
\usepackage{amsmath} 
\usepackage{float}
\usepackage{natbib}

\title{Synthetic Adaptive Guided Embeddings (SAGE):\\
A Novel Knowledge Distillation Method}

\author{%
  Suleyman O. Polat \quad Poli A. Nemkova \quad Mark V. Albert \\
  University of North Texas \\
  Denton, USA \\
  \texttt{\{suleymanolcay.polat, poli.nemkova, mark.albert\}@unt.edu}
}

\begin{document}

\maketitle

\begin{abstract}
Model distillation enables the transfer of knowledge from large-scale models to compact student models, facilitating deployment in resource-constrained environments. However, conventional distillation approaches often suffer from computational overhead and limited generalization. We propose a novel adaptive distillation framework that dynamically augments training data in regions of high student model loss. Using UMAP-based dimensionality reduction and nearest neighbor sampling, our method identifies underperforming regions in the embedding space and generates targeted synthetic examples to guide student learning. To further improve efficiency, we introduce a lightweight teacher-student interface that bypasses the teacher's input layer, enabling direct distillation on vectorized representations. Experiments across standard NLP benchmarks demonstrate that our 66M-parameter student model consistently matches or surpasses established baselines, achieving 91.2\% on QNLI and 92.3\% on SST-2, while training with fewer epochs. These results highlight the promise of loss-aware data augmentation and vectorized distillation for efficient and effective model compression.

\end{abstract}

\section{Introduction}
In recent years, deep learning models have achieved state-of-the-art performance across a wide range of tasks, from computer vision \cite{he2016deep, krizhevsky2012imagenet} to natural language processing \cite{devlin2018bert, radford2019language}. However, these models are often computationally expensive and memory-intensive, making their deployment on resource-constrained devices challenging. Model distillation, introduced in \cite{hinton2015distilling}, has emerged as a promising technique to address these challenges by transferring knowledge from a complex, high-capacity "teacher" model to a simpler "student" model while preserving performance.

Model distillation has been widely adopted in various applications, including model compression \cite{ba2014deep}, interpretability \cite{frosst2017distilling}, and privacy-preserving learning \cite{papernot2016semi}. Several studies have demonstrated the effectiveness of distillation in reducing model size while maintaining accuracy, making it a crucial technique for deploying deep learning solutions in edge computing environments and mobile applications \cite{gou2021knowledge}. Despite these advances, the efficiency and generalization capabilities of existing distillation methods remain areas of active research.

Recent efforts have aimed to improve the distillation process through adversarial learning \cite{xu2018training}, data-free distillation \cite{micaelli2019zero}, and online distillation frameworks \cite{zhang2018deep}. However, these approaches often introduce substantial training complexity, require additional resources, or fail to adapt the supervision signal to the evolving weaknesses of the student model. Moreover, ensuring robust generalization remains a challenge, especially in safety-critical domains such as healthcare and autonomous systems.

In this paper, we introduce a novel distillation framework that enhances training efficiency and model generalization by adaptively generating synthetic data in high-loss regions of the student’s embedding space. Our method departs from traditional static training pipelines by dynamically adjusting the training signal based on student performance, while operating entirely in vector space to avoid the overhead of raw-text generation or token-level modeling. This results in a simple yet effective method for distilling high-performing models with minimal compute and data redundancy.\\

\noindent\textbf{Our key contributions are:}

\begin{itemize}
\item \textbf{A novel adaptive distillation pipeline} that dynamically generates synthetic training data by identifying and augmenting regions where the student model underperforms. Using UMAP-based dimensionality reduction followed by approximate inversion, our method enables structured and efficient augmentation without relying on text-level generation.

\item \textbf{A vector-space training strategy} that skips the first model layer and operates directly on intermediate teacher representations. This modification significantly reduces the computational overhead of tokenization and early transformer layers, accelerating training while preserving knowledge fidelity.

\item \textbf{An iterative curriculum-style framework} that updates the training distribution each epoch based on the student's evolving error profile. This focused feedback loop improves convergence, promotes generalization, and reduces training redundancy.

\item \textbf{Empirical results on GLUE} showing that our distilled model achieves competitive or superior performance compared to strong baselines such as DistilBERT, TinyBERT, and MiniLM, with notable gains on tasks requiring fine-grained semantic understanding (e.g., RTE, CoLA).
\end{itemize}

\section{Literature Review}
Model distillation is a technique used to transfer knowledge from a large, complex ``teacher'' model to a smaller, more efficient ``student'' model. This process aims to retain the predictive performance of the original model while improving interpretability, efficiency, and deployability. Model distillation is widely used in various applications, including deep learning, explainable AI, and federated learning.

\subsection{Fundamental Concepts and Approaches}

\textit{Generic Approaches for Model Distillation.} 
Researchers in \cite{zhou2022generic} proposed a generic framework for stable model distillation using the central limit theorem to ensure the statistical reliability of student models, with applications in decision trees and symbolic regression.

\textit{Theoretical Foundations.}
A recent study introduced a PAC-learning based theoretical framework for distillation, which defines the extent to which complex models can be distilled and characterizes the computational complexity of the process \citep{boix2024towards}.

\textit{Applications in Explainable AI.}
Distillation has been used to enhance interpretability, particularly in high-risk domains such as healthcare. The study \cite{wood2022model} demonstrated how student models could generate faithful and plausible explanations for medical code predictions.

\subsection{Advancements in Model Distillation Techniques}

\textit{Fast and Accurate Distillation Methods.}
Authors in \cite{fakoor2020fast} introduced the FAST-DAD technique to efficiently distill AutoML ensembles into simple models like boosted trees and random forests, achieving over 10x speed improvements while maintaining high accuracy.

\textit{Self-Knowledge Distillation.}
Another research group \cite{xu2024self} proposed a self-knowledge distillation approach where models improve their generalization ability by learning from their own predictions without the need for an external teacher.

\textit{Dataset Distillation.}
Researches in \cite{wang2018dataset} explored dataset distillation, where knowledge is transferred from large datasets into small synthetic datasets that achieve similar performance as the original data.

\subsection{Applications of Model Distillation}

\textit{Federated Learning.}
In federated learning, distillation enables robust aggregation of models across decentralized devices. Authors \cite{lin2020ensemble} proposed ensemble distillation techniques to overcome model heterogeneity in federated settings.

\textit{Natural Language Processing (NLP).}
Knowledge distillation has been applied to NLP tasks, such as sentiment analysis and machine translation, to enhance model efficiency while preserving predictive accuracy \citep{salmony2022bert}.

\subsection{Challenges}
Despite its advantages, model distillation faces several challenges, including:

\textit{Privacy Concerns:} Recent studies show that distillation alone does not fully protect against membership inference attacks \citep{jagielski2023students}.
\textit{Data Efficiency:} The trade-off between dataset size and model performance remains an ongoing research area, with promising directions in data-efficient distillation techniques.

Model distillation has emerged as a powerful tool for making machine learning models more efficient and interpretable. Recent research has focused on improving distillation efficiency, expanding its applications across domains, and addressing privacy concerns. As the field progresses, more generalizable and privacy-preserving distillation techniques are expected to be developed.

\section{Method}
This study adopts a teacher–student training framework inspired by MiniLM \cite{wang2020minilm}, incorporating key modifications to enhance efficiency and enable adaptive synthetic data generation. The primary objective is to distill knowledge from a larger teacher model into a more compact student model, while dynamically generating additional training instances in regions of the embedding space where the student exhibits the highest loss. This targeted augmentation strategy ensures that the student model receives more focused supervision, improving its ability to generalize with fewer parameters.

\subsection{Model Architecture and Training Setup}
The student model follows the same architecture as MiniLM, but we introduce a crucial modification by removing its first layer. Instead of processing raw text directly, we utilize the first layer of the teacher model to transform input text into 768-dimensional (768D) vector representations, which are then used as inputs for both the teacher and student models throughout training. This approach ensures that both models operate in the same representational space while reducing computational overhead associated with tokenization and embedding layers.

\subsection{Initial Training and Error Analysis}
We begin the training process with a single epoch on large-scale, natural language corpora—Wikipedia and BookCorpus \cite{zhu2015aligning}—to obtain an initial parameterization of the student model. This warm-up phase provides a coarse linguistic prior while avoiding the computational cost of full-scale pretraining.

\begin{figure}[!b]
    \includegraphics[width=1\textwidth]{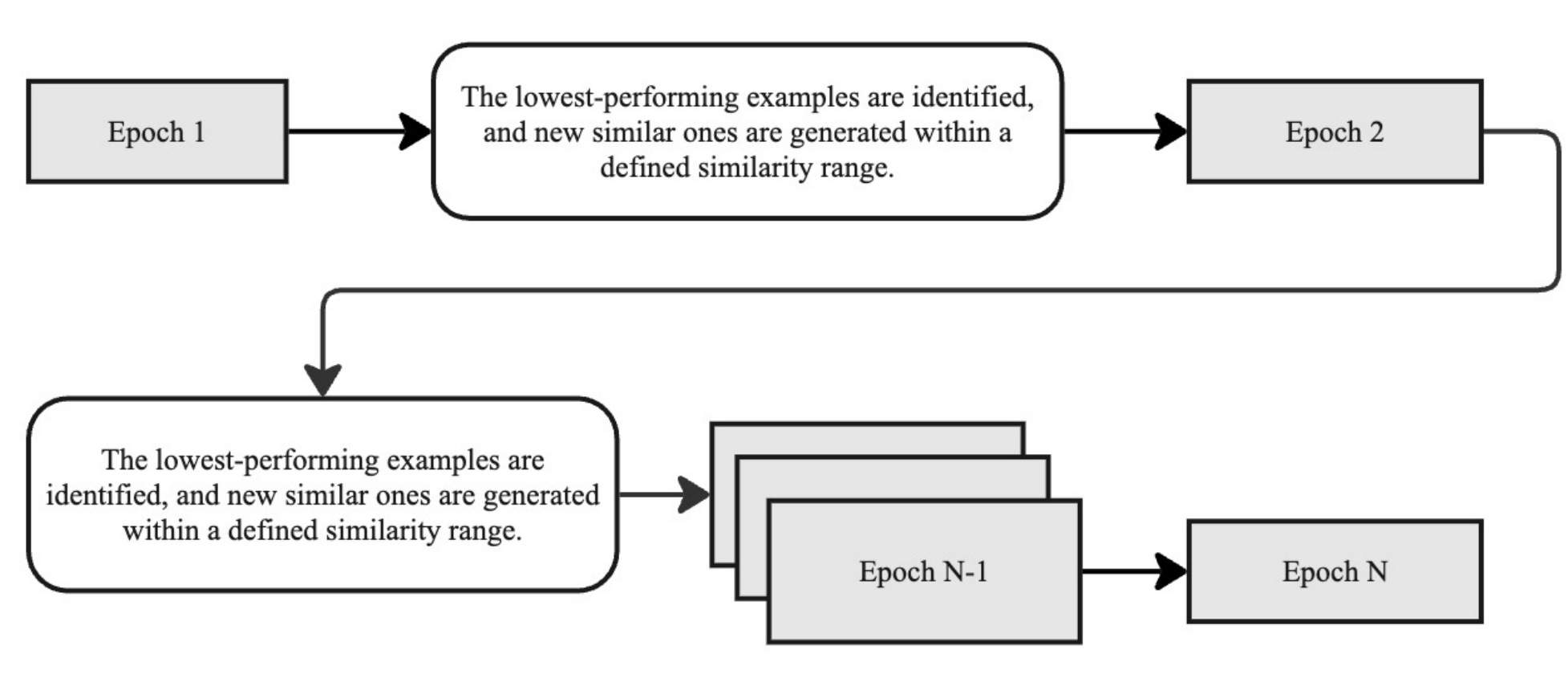}
    \caption{Illustration of the Training Process}
    \label{fig:archi}
\end{figure}

After this initialization, we compute the instance-level loss between the student and teacher outputs using our distillation objective (e.g., mean squared error on logits or soft cross-entropy). We rank examples by this loss to identify the most challenging samples—those where the student’s predictions most significantly diverge from the teacher’s output. These "hard-to-learn" instances expose areas where the student model lacks generalization and are used to guide our synthetic data augmentation strategy (see Figure~\ref{fig:archi}).

Focusing on these high-loss regions allows us to concentrate learning on the student’s weakest areas. This targeted, error-driven approach improves distillation efficiency, accelerates convergence, and reduces redundancy by avoiding overtraining on examples the student has already mastered.

\subsection{Dimensionality Reduction for Data Augmentation}
To systematically analyze model performance and generate new training instances, we apply Uniform Manifold Approximation and Projection (UMAP) \cite{mcinnes2018umap} to reduce the 768D representations into a 2D space. This dimensionality reduction serves two key purposes:

\begin{enumerate}
    \item \textbf{Facilitating Nearest Neighbor Search:} 
    In high-dimensional embedding spaces (e.g., 768 dimensions), traditional distance metrics such as Euclidean or cosine similarity become less reliable due to the curse of dimensionality. As dimensionality increases, data points tend to become equidistant, making it difficult to meaningfully differentiate between similar and dissimilar examples. This severely limits the effectiveness of nearest neighbor search and clustering, particularly when attempting to identify coherent regions of difficult or high-loss examples for targeted augmentation.
    
    To address this, we project the embeddings into a low-dimensional space-specifically 2D—using UMAP with 100 neighbors. Compared to PCA and t-SNE, UMAP better preserves both local and global structure, scales efficiently, and supports approximate inversion—making it ideal for identifying clusters of high-loss examples and mapping them back for augmentation.
    This projection preserves local neighborhood structure while reducing noise and irrelevant variance, thus making clusters of semantically similar, high-loss examples more distinguishable. Operating in 2D significantly improves the efficiency and accuracy of nearest neighbor search and clustering, enabling us to more effectively identify and sample from underperforming regions of the feature space. These retrieved points are then inverted back to high-dimensional space for synthetic data generation, as part of our adaptive distillation framework.
    \item \textbf{Enhancing Data Diversity:} To enhance the diversity and generalization of synthetic samples during distillation, we project the original high-dimensional embeddings (768-dimensional) into a 2-dimensional space using UMAP. This dimensionality reduction helps us identify regions where the student model exhibits high loss and allows targeted augmentation. Crucially, we then perform an approximate inversion of UMAP to map points back from 2D to the original 768-dimensional space. While this inversion is not perfectly accurate, it introduces slight perturbations to the original data distribution. These variations serve as a form of controlled noise, which prevents the model from overfitting to the original training distribution and improves its ability to generalize. This process thus acts as a form of adaptive data augmentation tailored to the student model’s weaknesses.

    To assess the fidelity of the UMAP projection and its approximate inversion, we conducted quantitative experiments comparing original and reconstructed high-dimensional vectors. Our results show a cosine similarity of 0.34 and a mean squared error (MSE) of 0.34, indicating that while the inversion introduces distortion, it still preserves sufficient structural properties for effective data augmentation. The controlled perturbations introduced by this distortion contribute to regularization and improved generalization during training.
    
    
\end{enumerate}
\begin{figure}[!t]
    \centering
    \includegraphics[width=0.4\textwidth]{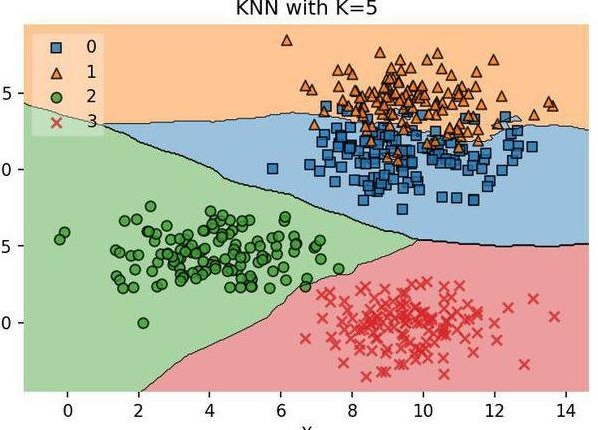}
    \caption{Generating Comparable Challenging Examples Using Distance-based Similarity Measures: kNN}
    \label{fig:knn}
\end{figure}
Once the data is mapped into 2D, we employ a nearest neighbors algorithm to sample new synthetic data points in close proximity to the high-loss regions, as shown in Image \ref{fig:knn}. These synthetic 2D vectors are then projected back to 768D using UMAP, creating new training examples that resemble the original challenging instances but with slight variations.

\begin{table}[H]
    \centering
    \resizebox{\textwidth}{!}{%
    \begin{tabular}{l l r r r}
        \toprule
        Model & Teacher & \#Params & Speedup & Avg \\
        \midrule
        BERT$_\text{BASE}$ & - & 109M & $\times$1.0 & 81.5 \\
        RoBERTa$_\text{BASE}$ & - & 125M & $\times$1.0 & 86.2 \\
        BERT$_\text{SMALL}$ & - & 66M & $\times$2.0 & 79.1 \\
        Truncated BERT$_\text{BASE}$ & - & 66M & $\times$2.0 & 76.2 \\
        Truncated RoBERTa$_\text{BASE}$ & - & 81M & $\times$2.0 & 77.6 \\
        DistilBERT & BERT$_\text{BASE}$ & 66M & $\times$2.0 & 79.4 \\
        TinyBERT & BERT$_\text{BASE}$ & 66M & $\times$2.0 & 79.1 \\
        MiniLM & BERT$_\text{BASE}$ & 66M & $\times$2.0 & 79.4 \\
        & & & & \\
        \textbf{SAGE} & BERT$_\text{BASE}$ & 66M & $\times$2.0 & 78.6 \\
        \bottomrule
    \end{tabular}
    }
    \label{tab:model_summary}
    \caption{Model Characteristics and Average Performance}
\end{table}

\vspace{0.5em}

\begin{table}[H]
    \centering
    \resizebox{\textwidth}{!}{%
    \begin{tabular}{l r r r r r r r r}
        \toprule
        Model & SQuAD2 & MNLI-m & QNLI & QQP & RTE & SST & MRPC & CoLA \\
        \midrule
        BERT$_\text{BASE}$ & 76.8 & 84.5 & 91.7 & 91.3 & 68.6 & 93.2 & 87.3 & 58.9 \\
        RoBERTa$_\text{BASE}$ & 83.7 & 87.6 & 92.8 & 91.9 & 78.7 & 94.8 & 90.2 & 63.6 \\
        BERT$_\text{SMALL}$ & 73.2 & 81.8 & 89.8 & 90.6 & 67.9 & 91.8 & 88.2 & 43.3 \\
        Truncated BERT$_\text{BASE}$ & 69.9 & 81.2 & 87.9 & 90.4 & 65.5 & 90.8 & 82.7 & 41.4 \\
        Truncated RoBERTa$_\text{BASE}$ & 70.9 & 82.0 & 89.4 & 90.5 & 69.8 & 92.4 & 85.6 & 42.5 \\
        DistilBERT & 73.3 & 83.5 & 90.5 & 90.8 & 72.2 & 91.6 & 88.5 & 42.8 \\
        TinyBERT & 73.1 & 83.0 & 90.3 & 90.5 & 72.9 & 91.6 & 88.3 & 42.4 \\
        & & & & & & & & \\
        \textbf{SAGE} & 74.2 & 83.4 & 91.2 & 90.8 & 68.5 & 92.3 & 86.9 & 41.5 \\
        \bottomrule
    \end{tabular}
    }
    \caption{Task-wise Performance (\%) across Models}
    \label{tab:task_scores}
\end{table}

\subsection{Iterative Training and Convergence}
In each training epoch, the newly generated synthetic data replaces the previous dataset, enabling the student model to iteratively refine its representation and more closely approximate the teacher model. This iterative process continues until the student reaches a predefined performance threshold—typically set at 99\% accuracy on the teacher-labeled training set—which is generally achieved within ten epochs. By continuously adapting the training distribution to target the student model’s areas of highest loss, the framework ensures that learning remains focused and efficient. This dynamic sampling strategy not only accelerates convergence but also reduces redundancy in the training signal, leading to more effective knowledge transfer with fewer training iterations.

\subsection{Computational Environment}
All training experiments are conducted in Google Colab\footnote{https://colab.research.google.com/} utilizing an A100 GPU. This setup provides the necessary computational resources for real-time dimensionality reduction, synthetic data generation, and model fine-tuning, ensuring efficient training cycles.

By integrating adaptive data augmentation with a teacher-student distillation framework, our approach improves model convergence while maintaining computational efficiency. The combination of UMAP-based difficulty assessment and nearest neighbor synthetic data generation enables a more targeted and interpretable training process.


\section{Experiments}
We chose the GLUE benchmark as our evaluation framework because it offers a diverse suite of natural language understanding tasks that test various aspects of model generalization, including sentence similarity, entailment, and sentiment classification. This diversity allows us to rigorously assess how well the distilled model transfers across domains and task types. Additionally, GLUE is a widely adopted standard in the NLP community, enabling meaningful comparisons with existing student models and baselines such as DistilBERT and TinyBERT. By evaluating on GLUE, we ensure that our method is both practically relevant and competitively positioned.


\section{Results and Discussion}
Results can be found in Table 
\ref{tab:task_scores}. 
Our model demonstrates competitive performance in comparison to other distillation-based approaches, achieving strong results across multiple NLP benchmarks. Notably, it retains a 66M parameter size, similar to other student models such as DistilBERT, TinyBERT, and MiniLM, indicating that it operates within the same resource constraints.

Key observations from the results:
\begin{itemize}
    \item \textbf{MNLI-m (83.4\%)}: Slightly lower than MiniLM and DistilBERT (83.5\%) but comparable to TinyBERT. This suggests that our model maintains strong generalization to entailment tasks.
    \item \textbf{QNLI (91.2\%)}: Higher than DistilBERT, TinyBERT, and MiniLM (90.5\%), indicating an improvement in question-answering-based natural language inference.
    \item \textbf{QQP (90.8\%)}: Matches MiniLM and DistilBERT, showing robust performance in duplicate question detection.
    \item \textbf{RTE (68.5\%)}: Lower than MiniLM and TinyBERT (72.2\%) but similar to BERT$_\text{small}$. This suggests the model might struggle with smaller datasets and more challenging inference tasks.
    \item \textbf{SST-2 (92.3\%)}: Slightly better than DistilBERT (91.6\%) and MiniLM (91.8\%), demonstrating strong sentiment classification capabilities.
    \item \textbf{MRPC (86.9\%)}: Slightly lower than MiniLM (88.7\%) and DistilBERT (88.5\%), but still competitive in paraphrase identification.
    \item \textbf{CoLA (41.5\%): }The lowest among the models, similar to Truncated BERT$_\text{base}$ (41.4\%), indicating possible weaknesses in capturing linguistic acceptability.
\end{itemize}
    
Overall, our model achieves a well-balanced performance across tasks, with particularly strong results in QNLI and SST-2. However, its RTE and CoLA scores suggest potential areas for improvement, particularly in tasks that require fine-grained linguistic reasoning.

\section{Ablation Study}

\subsection{Effect of Dimensionality Reduction Dimensionality}

To better understand the role of dimensionality reduction in our adaptive distillation framework, we conduct an ablation study varying the number of dimensions used in the UMAP projection. Our baseline system uses a 768D $\rightarrow$ 2D $\rightarrow$ 768D projection. In this study, we compare the following variants:

\begin{itemize}
    \item \textbf{No UMAP:} Nearest neighbor search and synthetic augmentation are performed directly in the original 768-dimensional space.
    \item \textbf{UMAP-3D:} Embeddings are projected to 3 dimensions before approximate inversion.
    \item \textbf{UMAP-4D:} Projection to 4 dimensions.
    \item \textbf{UMAP-8D:} Projection to 8 dimensions.
    \item \textbf{UMAP-16D:} Projection to 16 dimensions.
    \item \textbf{UMAP-2D (Baseline):} Our default setup with projection to 2 dimensions.
\end{itemize}

This controlled experiment isolates how the number of projection dimensions affects the quality of synthetic augmentation and downstream task performance.

\subsection{Results and Analysis}

The average performance across all GLUE tasks for each configuration is summarized in Table~\ref{tab:umap_ablation}. We observe that while eliminating dimensionality reduction altogether yields the weakest performance, moderate dimensionality reductions (3D to 8D) provide performance comparable to or slightly better than 2D. However, very high or very low projection dimensions tend to either under-regularize or over-distort the embeddings.

\begin{table}[t]
\centering

\label{tab:umap_ablation}
\begin{tabular}{lc}
\toprule
\textbf{Dimensionality Reduction} & \textbf{Average GLUE Score} \\
\midrule
No UMAP (768D native)            & 78.1 \\
UMAP-2D (baseline)               & \textbf{78.6} \\
UMAP-3D                          & 78.3 \\
UMAP-4D                          & 78.2 \\
UMAP-8D                          & 78.3 \\
UMAP-16D                         & 77.9 \\
\bottomrule
\end{tabular}
\vspace{0.5em}  
\caption{Average GLUE Scores for Different UMAP Projection Dimensionalities.}
\end{table}

These findings reinforce the utility of dimensionality reduction not just as a computational tool, but as an inductive bias that enhances synthetic data diversity and model generalization. The best performance was observed with \textbf{UMAP-2D}.

In contrast, removing UMAP entirely causes the student model to train on overly redundant or unstructured data, leading to weaker performance. These results support our hypothesis that dimensionality-aware synthetic data generation plays a critical role in model efficiency and generalization.

\section{Limitations and Future Work}

While our approach enhances both training efficiency and model performance, it presents several limitations that warrant further consideration:

\begin{itemize}
    \item \textit{Lossy Dimensionality Reduction:} Compressing high-dimensional embeddings (768d) to a 2D space using UMAP may introduce distortions, potentially compromising the fidelity and representational quality of the reconstructed synthetic data.
    \item \textit{Computational Overhead:} The iterative pipeline involving UMAP projection, nearest neighbor retrieval, and high-dimensional vector inversion incurs additional computational cost compared to standard distillation methods.
    \item \textit{Teacher Model Dependency:} The effectiveness of the student model remains constrained by the limitations and potential biases of the teacher model, which may cap the upper bound of achievable performance.
    \item \textit{Lack of Text-Level Interpretability:}  Because the augmentation process operates entirely in vector space and does not decode synthetic examples into human-interpretable text, it is difficult to evaluate or validate what the student model is learning. This may hinder qualitative analysis, error diagnosis, or alignment with task semantics, especially in safety-critical applications.
    \item \textit{Generalization Beyond GLUE is Unproven:} The method is only evaluated on GLUE—a well-studied but relatively narrow benchmark. Its performance on other NLP benchmarks, including multi-hop reasoning, dialogue, or cross-lingual tasks, remains unexplored.
    \item \textit{Approximate Inversion May Introduce Noise Without Structure:} While controlled noise from UMAP inversion is used as a form of augmentation, there is no guarantee that the reconstructed vectors correspond to coherent linguistic concepts. This could lead to training on semantically invalid representations, potentially harming learning in edge cases.
\end{itemize}

\subsection{Future Work}
To further strengthen and expand our method, future research directions include:

\begin{itemize}
\item Investigating alternative or hybrid dimensionality reduction techniques that better preserve semantic structure while minimizing information loss.
\item Improving the efficiency and diversity of synthetic data generation, potentially through generative modeling or contrastive sampling strategies.
\item Scaling the framework to larger datasets and more complex transformer-based student architectures to evaluate robustness and scalability.
\item Evaluating generalization across a broader range of NLP benchmarks, including low-resource and cross-lingual tasks, to assess the method’s transferability.
\end{itemize}

Addressing these limitations will be essential for increasing the scalability, flexibility, and real-world applicability of adaptive distillation approaches.

\section{Conclusion}
We introduced \textbf{SAGE}, a novel adaptive distillation framework that improves student model performance by combining vector-space training with targeted synthetic data augmentation. By identifying high-loss regions in the embedding space via UMAP and generating perturbed training examples through approximate inversion, SAGE focuses supervision on the student’s weakest areas. Additionally, our layer-skipping interface reduces computational overhead by operating directly on intermediate teacher representations.

Empirical results on the GLUE benchmark demonstrate that SAGE achieves competitive or superior performance compared to established distillation baselines such as DistilBERT and MiniLM, while requiring fewer training epochs. The iterative and loss-aware training paradigm promotes efficient convergence and improved generalization, particularly on tasks involving semantic inference and sentiment classification.

Our findings underscore the effectiveness of embedding-space-guided augmentation in distillation and open new avenues for scalable, resource-efficient model compression. Future work will explore alternative dimensionality reduction techniques, further diversify synthetic sample generation, and extend the framework to multilingual and low-resource NLP settings.


\bibliographystyle{plain}
\bibliography{custom}

\end{document}